# Lenient Evaluation of Japanese Speech Recognition: Modeling Naturally Occurring Spelling Inconsistency

**Shigeki Karita** and **Richard Sproat** and **Haruko Ishikawa**
Google DeepMind / Shibuya 3-21-3 Tokyo Japan
{karita,rws,ishikawa}@google.com

## Abstract

*Word error rate* (WER) and *character error rate* (CER) are standard metrics in Speech Recognition (ASR), but one problem has always been *alternative spellings*: If one's system transcribes *adviser* whereas the ground truth has *advisor*, this will count as an error even though the two spellings really represent the same word.

Japanese is notorious for "lacking orthography": most words can be spelled in multiple ways, presenting a problem for accurate ASR evaluation. In this paper we propose a new *lenient* evaluation metric as a more defensible CER measure for Japanese ASR. We create a lattice of plausible respellings of the reference transcription, using a combination of lexical resources, a Japanese text-processing system, and a neural machine translation model for reconstructing kanji from hiragana or katakana. In a manual evaluation, raters rated 95.4% of the proposed spelling variants as plausible. ASR results show that our method, which does not penalize the system for choosing a valid alternate spelling of a word, affords a 2.4%–3.1% absolute reduction in CER depending on the task.

## 1 Introduction: "Word" error rate

For decades, a standard measure of performance in Automatic Speech Recognition (ASR) has been *word error rate* (WER), which gives a measure of how poorly a transcription hypothesized by the ASR system matches a reference transcription and which, while often criticized—e.g. (Wang et al., 2003)—is still widely used. While the expression *WER* uses the term *word*, it is important to note that what is matched is not really words, but rather *spelled forms*. To take a simple example from English, the reference transcription might have the token *advisor*, whereas the corresponding token in the hypothesis is *adviser*. Although these are variant spellings of the same word, the system would be assessed as getting the word wrong, since the spellings do not match. If one used instead *character error rate* (CER), the effect of the spelling discrepancy would be of course be less, but there would still be an error. Arguably this should really not count as an error, since the spelling alternates are both valid.

*Orthographic variation* (Meletis and Dürscheid, 2022, Section 4.6), is common in the world's writing systems, but for many systems the effect is a minor one. In English, for example, orthographic variation is of two main types: regional variation, in particular British versus American spelling (e.g. *neighbour* vs. *neighbor*); and more or less free variation within a regional variety such as the *advisor/adviser* example above, or issues such as whether to write a space in noun compounds (e.g. *doghouse* vs. *dog house*). In the former case, one can argue that a spelling discrepancy should count as an error since in contexts where, say, *flavour* would be an appropriate spelling, *neighbor* would not be, and vice versa. In the latter case, the variants should probably not be counted as errors, but a naive WER or CER computation would so count them. Still, since the amount of such spelling variation is relatively small, one can usually ignore this effect, or use cleanup scripts to handle the few cases that occur. WER is a fiction, but it is a fiction that can largely be ignored.

## 2 Japanese spelling inconsistency

In Japanese, unlike in English, spelling variation is rampant, and the fiction becomes too great to be ignored. Japanese spelling is very inconsistent, with many words that have kanji (Chinese character) spellings also appearing in text in hiragana (one of the two syllabaries used in Japanese), or even, for emphasis or other reasons, in katakana (the other syllabary). Thus common words like だめ (hiragana) *dame* 'not allowed' also frequently appear as ダメ (katakana) for emphasis, but there is also a somewhat infrequent but nonetheless occurring form in kanji, 駄目. ください *kudasai* 'please' also frequently appears as 下さい. うまい *umai* 'good' can also be written as 上手い. If one's reference transcription has ダメ *dame* and the ASR system hypothesizes だめ, a naive WER/CER computation would count this as an error, even though these are both valid variant spellings.

There are many reasons for the variation. Some of them have to do with style—on which see Section 6. Katakana is frequently used to mark *emphasis* so that in Japanese, orthographic variation is used to mark what in English would involve either, to adopt the terminology of Meletis and Dürscheid (2022), *graphetic* variation such as italics, or *graphemic* variation such as capitalization. Joyce and Masuda (2019) give the example of *mechamecha* 'absurdly', normally written in katakana メチャメチャ, being written in kanji as 目茶目茶 in a sentence with foreign words or emphasized words, both written in katakana. They suggest the reason for the kanji spelling in this case is to provide visual distinctiveness. Spelling variants may also be used for artistic reasons (Lowy, 2021). One would like to have a measure of error rate that takes these sorts of variation into account.

One could of course propose developing reference transcriptions that are highly standardized so that, e.g., *dame* is always written だめ, thus eliminating the problem. Indeed corpora such as the Corpus of Spontaneous Japanese[1] exist that have highly standardized orthographic transcriptions. But this is not a practical solution in general for a couple of reasons. First, a large amount of potential training data that comes with transcriptions—for example YouTube videos—will not have been subjected to rigorous transcription guidelines, and the cost of retranscribing such data would be prohibitive. Second, downstream applications cannot be expected to adhere to whatever guidelines have been adopted, and one would like the flexibility to provide transcriptions that can match what downstream applications expect. Over and above this, normalizing everything to a standard spelling misses the fact that variation is a normal part of Japanese spelling, and one can ignore this only by adopting an artificial standard. We propose therefore to try to model what every native speaker/reader of Japanese knows, namely that だめ, ダメ and 駄目 are all legal ways to write *dame* 'not allowed'.

At the same time, one cannot allow the system to be too loose. To return to an example cited above, for うまい *umai* 'good', one can also have 上手い, as above, but in addition another possible written form is 美味い. The second spelling, 上手い just means 'good' (i.e, good at something), whereas 美味い means 'good tasting'. These two senses are both available for うまい, but they are not interchangeable, and this issue comes up if the ground truth has a kana spelling, whereas the hypothesized form uses kanji. If ground truth has うまい, and the system transcribes 美味い 'delicious', whereas a native speaker could tell from context that what was intended was 上手い 'good', this should count as an error. In reconstructing spelling variants in kanji from reference forms in kana, the system therefore needs to do sense disambiguation.

## 3 Proposed method

Our method starts with the creation of a lattice of possible respellings, given a reference transcription for an utterance. In order to illustrate the method, we consider the hypothetical reference transcription

| この | 拉麺 | は | うまい | 。 |
|------|------|-----|-------|---|
| *kono* | *rāmen* | *ha* | *umai* | . |

'this ramen is delicious'. The first step involves computing hiragana transcriptions for kanji sequences, which in the case at hand will yield らーめん for *rāmen*. In general tokens written in kanji may have multiple readings, but usually only one reading is appropriate for a given context. For this conversion we use a proprietary Japanese lattice-based text normalization system that uses a large dictionary, annotated corpora, rules, and linear classifiers to determine the most likely readings of kanji sequences in context. The system has a roughly 97% token accuracy on held out data. As is well-known, Japanese text lacks word separators, but one side-effect of the text normalization system is to produce a word-segmentation of the sentence. These word segments are used as the tokens for subsequent processing in our lattice construction.

For each hiragana word, we also want a katakana equivalent—cf., the example of だめ/ダメ above. This is a fairly straightforward conversion and in the example at hand would produce ラーメン for *rāmen*, which also happens to be the way this word is normally written.

This completes the conversion of kanji tokens into kana, and the next step is to convert in the other direction. For example, the last non-punctuation token in the utterance うまい *umai* 'delicious', also has a common kanji spelling 旨い. However as noted above, in this as in many other cases, one needs to be careful, since another possible spelling for うまい is 上手い, which would not be appropriate in this instance since it means 'skillful'. For this conversion we train a transformer-based neural machine translation model (NMT)—e.g. (Tay et al., 2020)—on Japanese web text where we first converted successive kanji spellings into hiragana using the text normalization system previously described. For example, consider the input sentence:

再び、MTサミットが日本で
*futatabi, MT samitto-ga nihon de*
'Again, the MT Summit is in Japan'

which contains two words containing kanji 再び *futatabi* 'again' and 日本 *nihon* 'Japan'. Consider the second of these, which has the hiragana transcription にほん. We replace this into the sentence above and tag it with a special tag `<to_kanji>...</to_kanji>` so that the input appears as

再び、　　MTサミットが　`<to_kanji>`　にほん

`</to_kanji>` で

and we train the NMT system to predict 日本, given this context. We also need to train the model to predict cases where kana spellings should not be replaced by kanji: For example the final で does not have a kanji variant, and so in this case we would produce a variant of the input sentence with that token tagged with `<to_kanji>...</to_kanji>`, and the system would be trained to replace it with itself.

The NMT transformer is configured with 6 layers, 8 attention heads and a hidden-layer dimension of 2048, and trained on a web corpus of 7.3 billion tokens. At runtime the model is applied to each hiragana word in the sentence in turn to predict whether that token should be replaced with a word spelled in kanji, and if so which word. The token error rate for *kanji restoration* on a held out corpus is approximately 3.8%. In the case at hand, the system would correctly predict 旨い as an appropriate spelling of うまい.

Finally, 旨い has another possible kanji spelling, 美味い, which we have already seen in Section 2. To allow for these variants we use lexical resources licensed from the CJK Institute (`www.cjki.org`), in particular the Japanese Orthographic Dictionary, which lists spelling equivalence classes for several tens of thousands of Japanese words.[2] These equivalence classes are 'safe' in the sense that one can substitute any spelling in the class for any other without considering the context of the word. Thus 美味い and 旨い both mean 'delicious' and can be safely substituted for each other. To the CJKI institute data we have added additional equivalence classes mined from various data sources such as Wikipedia, for a total of about 54,200 equivalence classes.

Figure 1 shows the complete lattice that is reconstructed for the input sentence given above.

Lattices are implemented in OpenFst (Riley et al., 2009), with weights represented in the

---

[2]Public resources such as JMDict (Breen, 2004) also contain some information on spelling variants. However, as we note in the Limitations section, unlike the JOD, JMDict has not been curated with a view to marking which spellings are interchangeable. Nonetheless, we plan to investigate incorporating additional data from JMDict in future work.

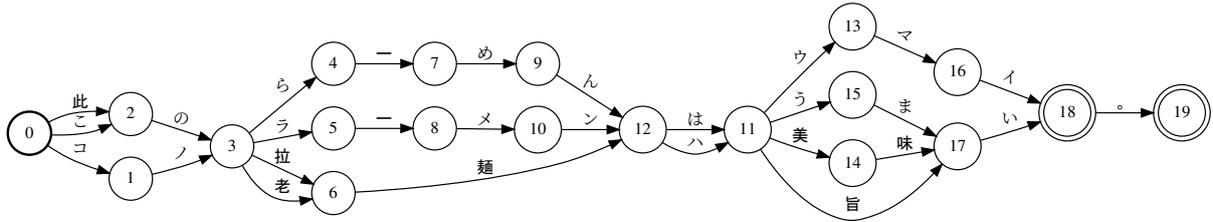

Figure 1: Final lattice computed for the reference transcription この拉麺はうまい。 'this ramen is delicious'.

Tropical semiring. During evaluation, the Levenshtein edit distance (Levenshtein, 1966) between the reference lattice and the hypothesized transcription is computed using the algorithm reported in Gorman and Sproat (2021), pp. 93–96.

As with standard CER, we define our lenient CER as the lattice edit distance—the sum of the substitution, insertion and deletion errors—divided by the number of characters in the best matching path in the reference lattice.

In future work (Section 6) we also wish to incorporate style/register language models to rank different transcriptions, and we will thus want to preserve language model weights for the various spelling alternatives. To that end, we first convert the Tropical weights into a <Tropical, Tropical> Lexicographic semiring (Sproat et al., 2014), where the first dimension is reserved for the edit distance weights, and the second dimension preserves the language model weights. This will guarantee that the path in the lattice closest to the hypothesized string is selected, with the language model score of that path preserved in the second dimension. After the shortest path has been computed, the result can be converted back to the Tropical semiring with just the (second-dimension) language model weights.

In the experiments reported in Section 5, we compare the results with multiple lattice variants, which are indicated with the terms boldfaced below:

1. The raw ground-truth transcription, represented as a trivial (single-path) lattice.

2. The lattice in (1) augmented with kana conversion via the text-normalization system (**+kana**).

3. The lattice in (2) augmented with the kanji restoration NMT model (**+kanji**).

4. The lattice in (3) augmented with the spelling equivalence classes (**+lexicon**).

## 4 Related Work

While the contribution of spelling variation to error rate computation for Japanese ASR has been noted—see Mishima et al. (2020), page 72—as far as we can tell, there has been no prior work that specifically addresses solutions to this problem. However, the problem of spelling variation in Japanese is similar to cases in other languages where no standardized spelling exists. For example, Ali et al. (2017)—and see also Ali et al., 2019)—present an approach for ASR for Arabic dialects. Unlike Modern Standard Arabic, which has an official and standardized orthography, Arabic regional varieties such as Levantine, Gulf Arabic, or Maghrebi are spoken languages that have no generally agreed standard written form. Nonetheless, particularly with the advent of social media, people increasingly communicate in Arabic dialects in written form. But since there is no prescribed standard there is a substantial amount of variation in how words are spelled. Ali et al. (2017) propose the *WERd* ("word error rate for dialects") metric, which depends on a spelling variants table, which they construct from social media. Variants are collected by mining tokens that share the same context, occur a sufficient number of times, and are within a Levenshtein edit-distance bound of each other. This kind of approach for finding potentially intersubstitutable terms has been used in other applications: for example, Roark and Sproat (2014) propose a similar approach for finding potential pairs of words and novel abbreviations of those words. Once the spelling variants table is constructed, Ali et al. (2017) use it to match ASR candidates against the reference transcription similar to the way in which our lattice-based matching works. Related work includes Nigmatulina et al. (2020), who report on an ASR system for Swiss German, which like dialectal Arabic, has no standard orthography, but where spellings are loosely based on pronunciation.

Another case of spelling variation can be found with transliteration, say when someone whose native language is Hindi using the Devanagari script, transliterates a Hindi word into English. As Roark et al. (2020) discuss, this problem has a practical application, since while keyboards for Devanagari and other South Asian scripts exist, they tend to be difficult to use, whereas many users are used to typing in English. Therefore many users prefer to type in Latin script transliteration, and have the system automatically convert to the native script. But this introduces a problem since, while there are standards for transliteration of South Asian languages into Latin script, few people adhere to them. The result is that one can find quite a large amount of variation in how to spell words in Latin script, whereas there is generally one way to correctly write a given word in the native script. Roark et al. (2020) investigated a variety of methods including both neural and pair n-gram methods, and found that they got the best performance with a pair 6-gram model using a Katz-smoothed trigram language model for the output.

While the above cases are similar to the problem with Japanese spelling variation, there is also an important difference. For dialectal Arabic, and transliterated South Asian languages, there is no standard, and so long as the message can be communicated, users are more or less unconstrained in how they will spell words. In the case of Japanese, spelling variation is not completely unconstrained: there are definitely *wrong* spellings for words, even if there is in any given case no *single* right spelling. While this does not dictate a particular approach to the problem, it does mean that the variation needs to be constrained by lexical knowledge implemented in some fashion.

Our use of Neural MT models for kanji restoration is related to the similar use of NMT

models for transliteration: see, e.g., Grundkiewicz and Heafield (2018) and Kundu et al. (2018).

Finally, we note that the problem of lenient evaluation comes up in other domains, for example in evaluation of MT systems. For example, Bouamor et al. (2014) argue that the rich morphology of Arabic has a negative impact on BLEU scores in that a naive application of BLEU can rank correct translations lower than incorrect ones. They propose a lenient metric they term "AL-BLEU", which takes morphological variation into account. They argue that this metric provides a more defensible evaluation metric.

# 5 Experiments

We investigated our proposed evaluation metrics on several Japanese ASR tasks. Using large-scale multiple domain datasets, we calculated error reductions from conventional naive CER, using lattices that incorporate the additional resources discussed in the last section. We also conducted human evaluations to validate the generated spelling alternatives.

## 5.1 ASR datasets

We evaluated on proprietary Japanese datasets in three domains: *Farfield*, *Voice-Search* (VS), and *YouTube* (YT)— respectively, domains involving far-field speech, voice search, and YouTube video segments, (Narayanan et al., 2019). These datasets contain anonymized and hand-transcribed utterances. The numbers of evaluated utterances were 15,693 (161,174 characters) for Farfield; 9,440 (78,606 characters) for VS; and 17,780 (238,662 characters) for YT.

## 5.2 ASR models

Our Japanese ASR models are Conformer-based Recurrent Neural Network Transducers (RNN-T) (Gulati et al., 2020). For the YT domain evaluation, we trained the ASR model only with a YT training set of 2,000 hours using a 17-layer, 512-dimensional, 8-attention-head, non-causal encoder, and a 5000-class character vocabulary. Apart from YT, our model was trained with all the multi-domain training sets of 25,000 hours using a 12-layer,

1024-dimensional, 8-attention-head, causal encoder, with a 6400-class *wordpiece* model vocabulary (Schuster and Nakajima, 2012).

## 5.3 Results with ASR tasks

Table 1 shows conventional WER and CER using the raw ground truth text, and CERs using our proposed target lattices for each ASR domain evaluation, as discussed in Section 3. In addition to the average error rates, we also computed ±95% confidence interval following (Vilar, 2008).

First, comparing WER and CER with the raw reference text, CERs were always lower than WERs in every domain. This is largely because WER depends on word boundaries estimated by a word segmenter, which can often lead to artificial mismatches between reference and transcription. CER, obviously, does not require word segmentation. For this reason, we evaluated our evaluation method by comparing baseline CER rates against the lenient CER lattice-based scoring.

When we added alternative kana spellings into the reference lattice (**+kana**), CERs were decreased by at least 2% absolute for all domains. More spellings from the kanji-restoration NMT (**+kanji**) and the lexicon (**+lexicon**) further reduced CERs to 2.4%–3.1% absolute depending on the domain. For example, VS was the most impacted domain, with a 25.16% relative error reduction.

A manual examination of cases of mismatch between the reference transcription and the hypothesized transcription in YT revealed many cases where one had kanji spellings and the other kana, or where one had hiragana and the other katakana, as one would expect given the discussion in Section 2. For example, the following pairs show, (1) kana-kana, (2) kana-kanji, (3) kanji-kana, and (4) kanji-kanji (false) errors between the ASR hypothesis (hyp) and reference ground truth (ref). Also given are romaji and a translation:

1.
hyp: イナバのチュールかな
ref: いなばのちゅーるかな
rom: *inaba no chūru-ka-na*
tra: Inaba Churu (dog treats)...

2.
hyp: 皆さんご機嫌よう
ref: みなさんごきげんよう
rom: *minasan gokigenyō*
tra: hello everyone

3.
hyp: がんばれ
ref: 頑張れ
rom: *ganbare*
tra: do your best

4.
hyp: 柔らかい設定になってます
ref: 軟らかい設定になってます
rom: *yawarakai settei ni nattemasu*
tra: it has a soft setting

The proposed system correctly produces these alternative spellings in the lattices created from the reference ground truth.

## 5.4 Manual evaluation of spelling variants

We evaluated spelling variants of 913 phrases generated by the model against the original spellings from the YT domain. More specifically, we evaluated phrases extracted from the transcription in the reference lattice that had the best edit-distance score when matched against the ASR hypothesis. The phrases varied in length from 1 to 93 characters, where 80% are 2 to 15 characters. Three trained raters were asked to assign each variant to one of the following bins:

- **Acceptable**: Spelling variants are without errors and acceptable.

- **Acceptable (Depending on the context)**: Acceptable in the context of a given phrase.

- **Great**: Great or better than the original spellings. Note that 'great' means that the selected alternate spellings are indeed commonly used valid spellings for the intended term, and 'better' means that the alternative spelling is even more natural, and easier and clearer for native speakers to read.

- **Great (Depending on the context)**: Great in the context of a given phrase.

- **Wrong**: Spelling contains errors.

|          | **Farfield**      | **VoiceSearch**   | **YouTube**       |
|----------|-------------------|-------------------|-------------------|
| WER      | $12.74 \pm 0.49$  | $11.58 \pm 0.52$  | $22.58 \pm 0.42$  |
| CER      | $12.53 \pm 0.49$  | $9.62 \pm 0.47$   | $18.36 \pm 0.38$  |
| **+kana** | $9.93 \pm 0.42$  | $7.68 \pm 0.40$   | $16.67 \pm 0.35$  |
| **+kanji** | $9.59 \pm 0.42$ | $7.45 \pm 0.39$   | $16.18 \pm 0.35$  |
| **+lexicon** | $9.46 \pm 0.41$ | $7.20 \pm 0.39$ | $15.84 \pm 0.35$  |

Table 1: Multi-domain ASR evaluation results with $\pm 95\%$ confidence interval. For "WER" and "CER", we used raw ground truth texts as reference targets. Other results show CERs using additional reference lattices augmented with alternative spellings of **+kana**, (kana)**+kanji**, and (kana+kanji)**+lexicon**, respectively. See Section 3 for details on what each of these augmentations means.

- **Wrong (Depending on the context)**: Spellings are inappropriate and considered as errors in the context of a given phrase.

Consolidated results show that over 95.4% of spelling variants are valid, and 16% are great or better than the original transcripts.

## 6 Conclusions and Future Work

In this paper we have proposed a lattice-based lenient evaluation method applied to computing character error rate in Japanese ASR. The method combines lexical resources, a Japanese text-processing system, and a neural MT system to reconstruct kanji from kana spellings in context. We evaluated on three different commercial Japanese ASR domains, and demonstrated a 2.4%–3.1% absolute reduction of CER—translating into an over 25% relative error reduction for the Voice Search domain.

Obviously these reductions in CER are not due to any improvement in the ASR method itself, but rather reflect a more defensible measure than naive comparison to a single reference transcription. This in turn points to the importance of taking spelling variation into account when evaluating systems on languages where such variation is simply a fact of life.

As noted in Section 3, we plan in future work to address another issue, namely style and register. While it is true that one often sees spelling variation for words even within the same text, it is also the case that style and register are important factors in deciding which spellings are felicitous in any given context. Thus while the word *kawaii* 'cute', has a kanji spelling 可愛い, that spelling would not usually be found in social media where the hiragana かわいい or katakana カワイイ variants would be more expected, especially if the goal is to communicate a more "friendly" message. We are currently experimenting with using style/register language models that are trained on different genres of text ranging from (informal) social media texts scraped from the Web to (formal) official Japanese government documents. In the context of the system described in this paper, the language models will be used to rank alternative spellings. Thus, a hypothesized spelling for a sentence may be a technically valid variant for a given reference transcription, but may also not be the most consistent in terms of style, and thus should be evaluated somewhat worse than a transcription that is more consistent. This would involve not only considering the edit distance measure—first dimension of the lexicographic semiring described in Section 3, but also the language-model cost in the second dimension.

We are also investigating using the spelling-variant-augmented reference lattices during training of the ASR system rather than just evaluation. Currently the ASR systems are trained with a single ground truth, which means that the ASR systems themselves are not sensitive to spelling variation. To this end, we are developing lattice-based loss functions that can be used during ASR training.

In addition to the above, we will continue to improve the system in various ways. We plan to include more lexical resources, such as publicly available resources like JMDict (see footnote 2), as well as improve the NMT-based kanji restoration model, with a goal to reducing human-judged unacceptable substitutions below the current 4.6%. An open question is whether large language models such as GPT-4 or Bard can be induced to provide judgments on whether two Japanese spellings are inter-substitable in a given context, and we also plan to investigate this in future work.

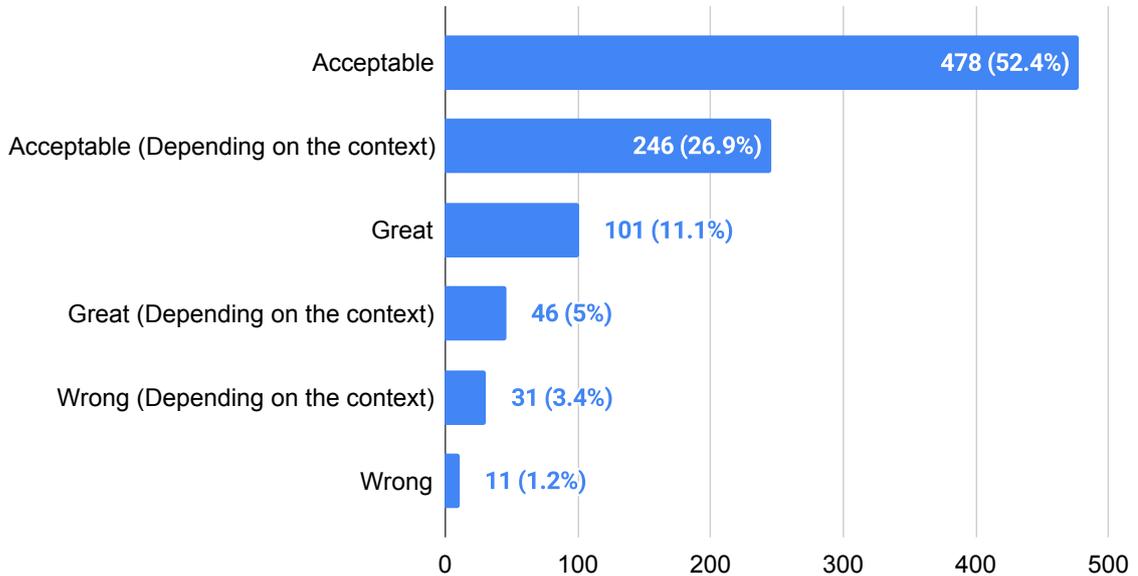

Figure 2: Manual evaluation results on spelling variants quality of 913 phrase pairs.

Finally, while Japanese provides a particularly rich example of spelling variation compared to other modern writing systems, as discussed in Section 4, there are many languages that are primarily oral, and for which there no accepted written standard. In such languages, one can expect a fair amount of variation in spelling when people attempt to write them, and the methods proposed in this paper could be applicable to such cases.

**Limitations**

Our work focuses on the problem of spelling variation in Japanese. The Japanese writing system is the most complex of any modern writing system (to find anything of comparable complexity, one would have to go back to cuneiform Akkadian or Hittite) and presents a unique range of issues that impact speech and language technology, one of which is the spelling variation discussed in this paper.

Nonetheless, as also noted in Section 6, we believe that the approach here should be applicable, perhaps with less dramatic results, to other cases where spelling variation occurs. This may be particularly an issue in languages that do not have a standardized writing system—e.g. Colloquial Arabic dialects—and where a large amount of spelling variation is often observed. However we have not evaluated the approach on this sort of data.

Our evaluation system is not open-sourced due to the propriety lexical resources, text normalizer and kana/kanji translators. The text normalizer could probably be replaced with, e.g., the open-source Mecab (Kudo, 2006) system, though we expect that performance would be degraded. Similarly our lexical resources could potentially be replaced with publicly available Japanese dictionaries such as JMDict (Breen, 2004), but again performance would probably suffer. Note in particular that unlike CJKI's Japanese Orthographic Dictionary, JMDict entries have not been carefully curated to indicate which spellings are interchangeable, and which are, rather, words with the same reading but distinct meanings. An informal manual evaluation we performed on potential spelling variant pairs that were extracted from JMDict entries nominally representing the same word sense, revealed that about 92% were valid variant spellings, but that the rest were either wrong, or at least unclear.

## Ethics Statement

The work reported in this paper relates to the impact of Japanese spelling inconsistency on the development and evaluation of Automatic Speech Recognition systems. The data used for our experiments is from a variety of sources and includes data from users, but it contains no Personal Identifiable Information. While it is possible that some of the data (especially data from YouTube) includes content that may have ethical concerns (e.g. hate speech, hurtful terminology, intentional or unintentional bias), the algorithms presented here are neutral with respect to these issues.

As discussed in Section 5.4, a subset of data was manually verified by human raters, all of whom were paid linguistic consultants hired through a third-party vendor.

## Acknowledgments

We thank our colleagues, in particular Yuma Koizumi, Llion Jones and Michiel Bacchiani for discussion and feedback. We also thank three reviewers for useful comments.